# A volumetric change detection framework using UAV oblique photogrammetry - A case study of ultra-high-resolution monitoring of progressive building collapse

Ningli Xu[a, d], Debao Huang[a,d], Shuang Song[a,c], Xiao Ling[a,c], Chris Strasbaugh[f], Alper Yilmaz[b,c], Halil Sezen[c] and Rongjun Qin[a,c,d,e]*

[a]*Geospatial Data Analytics Lab, The Ohio State University, Columbus, USA;*

[b]*Photogrammetric Computer Vision Lab, The Ohio State University, Columbus, USA*

[c]*Department of Civil, Environmental and Geodetic Engineering, The Ohio State University, Columbus, USA;*

[d]*Department of Electrical and Computer Engineering, The Ohio State University, Columbus, USA;*

[e]*Translational Data Analytics Institute, The Ohio State University, Columbus, USA;*

[f]*Engineering Technology Services, The Ohio State University, Columbus, USA*

* *Corresponding author: Rongjun Qin, qin.324@osu.edu, 2036 Neil Avenue, Columbus, Ohio, 43210, USA.*

In this paper, we present a case study that performs an unmanned aerial vehicle (UAV) based fine-scale 3D change detection and monitoring of progressive collapse performance of a building during a demolition event. Multi-temporal oblique photogrammetry images are collected with 3D point clouds generated at different stages of the demolition. The geometric accuracy of the generated point clouds has been evaluated against both airborne and terrestrial LiDAR point clouds, achieving an average distance of 12 cm and 16 cm for roof and façade respectively. We propose a hierarchical volumetric change detection framework that unifies multi-temporal UAV images for pose estimation (free of ground control points), reconstruction, and a coarse-to-fine 3D density change analysis. This work has provided a solution capable of addressing change detection on full 3D time-series datasets where dramatic scene content changes are presented progressively. Our change detection results on the building demolition event have been evaluated against the manually marked ground-truth changes and have achieved an F-1 score varying from 0.78 to 0.92, with consistently high precision ($0.92 - 0.99$). Volumetric changes through the demolition progress are derived from change detection and have shown to favorably reflect the qualitative and quantitative building demolition progression.

# 1. Introduction

Unmanned Aerial Vehicles have found their ever-expanding use in a wide range of applications through photogrammetric reconstruction and automated change detection (Yao, Qin, and Chen 2019; Colomina and Molina 2014). Given the relatively large relief difference with respect to the image scale, image-based change detection as used in typical remote sensing satellite image-based change detection (Tewkesbury et al. 2015) is no longer suitable for accurately identifying, locating, and delineating changes of objects of interest (Qin, Tian, and Reinartz 2016). Most of the existing work considered performing fine-grained change detection by utilizing standard photogrammetric products (i.e. digital surface models (DSM) and orthophoto) for various applications that range from crop growth analysis (Berni et al. 2009), illegal construction detection and waste disposals (Yoo et al. 2017). However, the DSM is still regarded as a 2.5 dimensional (2.5D) product such that the detected changes are often represented as height differences among multi-temporal datasets, which needs to be accurately registered either through ground control points (GCP) or reference tie points.

A greater advantage of oblique photogrammetric data is its ability to capture full three dimensional (3D) objects, such that the change detection can be performed from a volumetric perspective to understand complex changes in fine scale. Although manifested as advantageous, there are much fewer applications using 3D time-series datasets from UAV oblique images for change detection reported than those based on 2D or 2.5D datasets. In this paper, a case study is performed that utilizes UAV-oblique photogrammetry derived 3D data, to track a building demolition process by quantitatively estimating the volumetric changes in full 3D. For this purpose, we developed a fully automated change detection framework that performs a coarse-to-fine 3D density change detection. To be more specific, our approach requires no ground control points or manually identified tie points for model registration, and the proposed coarse-to-fine strategy through the temporal 3D dataset leverages well the noises and metric changes and is able to fully utilize the temporal coherence along the time-series 3D data.

This research has led to two unique contributions: 1) we demonstrate through a full-scale case study involving progressive building demolition event that was rarely reported through UAV-based monitoring, which UAV-oblique photogrammetric 3D data can be particularly useful to track fine-grained building damage progression or demolition; 2) we proposed an end-to-end workflow, with a fully automated coarse-to-fine 3D volumetric change detection method that advance the typical use of standard 2.5D geometric information to full 3D. Although our case study focuses on a building progressive demolition event, the proposed method can be readily used for any type of similar event for fine-grained change analysis of multi-temporal 3D data.

The rest of this paper is organized as follows: Section 2 describes existing studies closely related to our work; Section 3 introduces proposed change detection framework, with subsections describing details of the proposed approach. In Section 4, the case study and the collected data, experiments, and result analysis are described in detail. Section 5 concludes the paper by suggesting future improvements.

## 2.Related Work

Applications using UAVs for inspection, detection, and surveillances are highly disparate

(Gruen and Beyer 2001), and our application involves the use of UAV for collecting and processing multi-temporal 3D datasets, thus we briefly review related literature in the following two relevant aspects: 1) 3D dataset registration, and 2) progressive change detection using multi-temporal or time-series 3D dataset.

## 2.1 Multi-temporal UAV data registration

UAV images for 3D applications are usually captured following photogrammetric blocks or certain convergence patterns, i.e., (convergence image collection is centered at an object of interest). These overlapping images are mostly tagged with Global Positioning System (GPS) and (optionally) Inertial Measurement Unit (IMU) records that provide approximated geo-locational information and camera poses. Therefore, if the GPS and/or IMU information is used, the resulting 3D photogrammetric product (i.e. point clouds or 3D meshes) are considered to be approximately registered with a derivation on the scale of 10-15 meters depending on the accuracy of the initial GPS data (Turner, Lucieer, and Wallace 2014). These often constitute good starting points for explicit 3D registration algorithms such as Iterative Closest Points (ICP) or its variants (Rusinkiewicz and Levoy 2001; Zhang, Yao, and Deng 2021) for precise 3D registration, since ICP algorithms require dense 3D correspondences and iteratively estimate rigid or similarity transformation between 3D datasets.

In occasions without initial poses (e.g. no GPS and/or IMU records), existing approaches require at least three corresponding points to start the algorithms. However, such registration algorithms may be impacted by partial occlusions and outliers (or changes) when applied to UAV 3D datasets (Yang, Shi, and Carlone 2021). These type of registration algorithms are developed to register close-interval 3D scans and generally assume good content overlap and small outlier rates. Thus, they are not suitable for applications where dramatic scene changes are observed between 3D datasets. Another line of method focuses on refining the pose estimation of the multi-temporal photogrammetric images, such that the estimated poses across multi-temporal acquisitions sit accurately in the same datum, from which the derived 3D datasets are geometrically registered at the same level of accuracy as the pose estimation, leading to implicit 3D registration. A common strategy is to use GCPs that are assumed to be unchanged throughout the project, and separately geo-referencing each UAV collection using these GCPs (Altuntas 2019), which requires manual measurements in the image space. A few studies have identified that these manual efforts can be dramatically reduced by utilizing the large number of feature points that tie the datum of all collections to the datum of one dataset (Qin 2014) or just simply place all images into a common bundle adjustment (BA) (Li et al. 2017), herewith such an approach is denoted as feature-based 3D implicit registration. As a result of successful BA, the resulting 3D datasets separately from each collection with the refined poses can be precisely registered. This is regarded as a preferred approach since it does not demand good initial poses or 3D correspondences as the feature extraction in the image level is agnostic and can largely and automatically exploit unchanged regions through image content from multiple views, which can work under changes of the scene. However, potential drawbacks of this type of methods are two folds: first, the feature point extraction across different dates can be sensitive to illuminations and reflections (e.g., due to weather, rains, etc.). The chances of such happening grow with the number of multi-temporal collections; Second, the paradigm proposed by (Qin 2014) using one collection as the reference datum, or proposed by (Li et al. 2017) using all multi-date collections in one single BA, can be challenged by the growing scene changes resulting in growing outlier rates and consequently failure of the BA.

## 2.2 Time-series analysis and change detection in 3D

There are a plethora of change detection works using various types of remote sensing data and the use of UAV images is simply regarded as an extension of them (Yao, Qin, and Chen 2019; Hecheltjen, Thonfeld, and Menz 2014). Given that UAV images are often collected through a photogrammetric acquisition paradigm with potential derivable 3D data, here we focus on relevant work primarily in 3D. Most of the work characterized applications through bi-temporal 3D change detection, in which height differencing (Altuntas 2019), voxel-based, view-based, or segment-based methods (T. Pollard and Mundy 2007; Qin and Gruen 2014; Furukawa et al. 2020) were used to identify the presence of changes between 3D datasets, with post-refinement methods to improve the resulting detection, among which application using oblique datasets are less reported, as well as those using time-series 3D datasets. Bi-temporal 3D change detection often utilize voxel or volumetric presentations for change detection, and integrate sensor-specific properties, for example, Hebel et al. (Hebel, Arens, and Stilla 2013) characterized change detection by identifying conflict of evidence between airborne LiDAR (Light Detection and Ranging) data and the reference data in the laser pulse propagation path for a voxel-based occupancy grid; Finer change detection methods focusing on deformation modeling can be found in (Mukupa et al. 2017), under which the application is limited to small objects and building-façade and construction element levels (Kang and Lu 2011). A few works are focusing on 3D multi-temporal analysis at wider regions using satellite-derived 3D time-series data or DSMs, for example, Tian et al (Tian et al. 2016) used time-series DSMs generated from satellite stereos to enhance the build construction/demolition dynamics across a few years, in which the time-series data added a continuity constraint (i.e., building demolition and construction at the same location do not occur at high frequency), to improve the robustness of the change detection results, and similar works can be found in (Tian, Dezert, and Qin 2018). A notable line of work, instead of modeling the changes and geometric measures, models the probability of voxels in occupancy grid based on the radiometric consistencies of time-series images taken from multiple views (T. B. Pollard et al. 2010; Ulusoy and Mundy 2014), which can take single image or views from a different time as the input for change detection. However, since each image is tested against the radiometry distributions in a 3D volumetric grid, the changes are essentially modeled in 2D and do not possess information for 3D volume change extraction.

## 3. Methodology

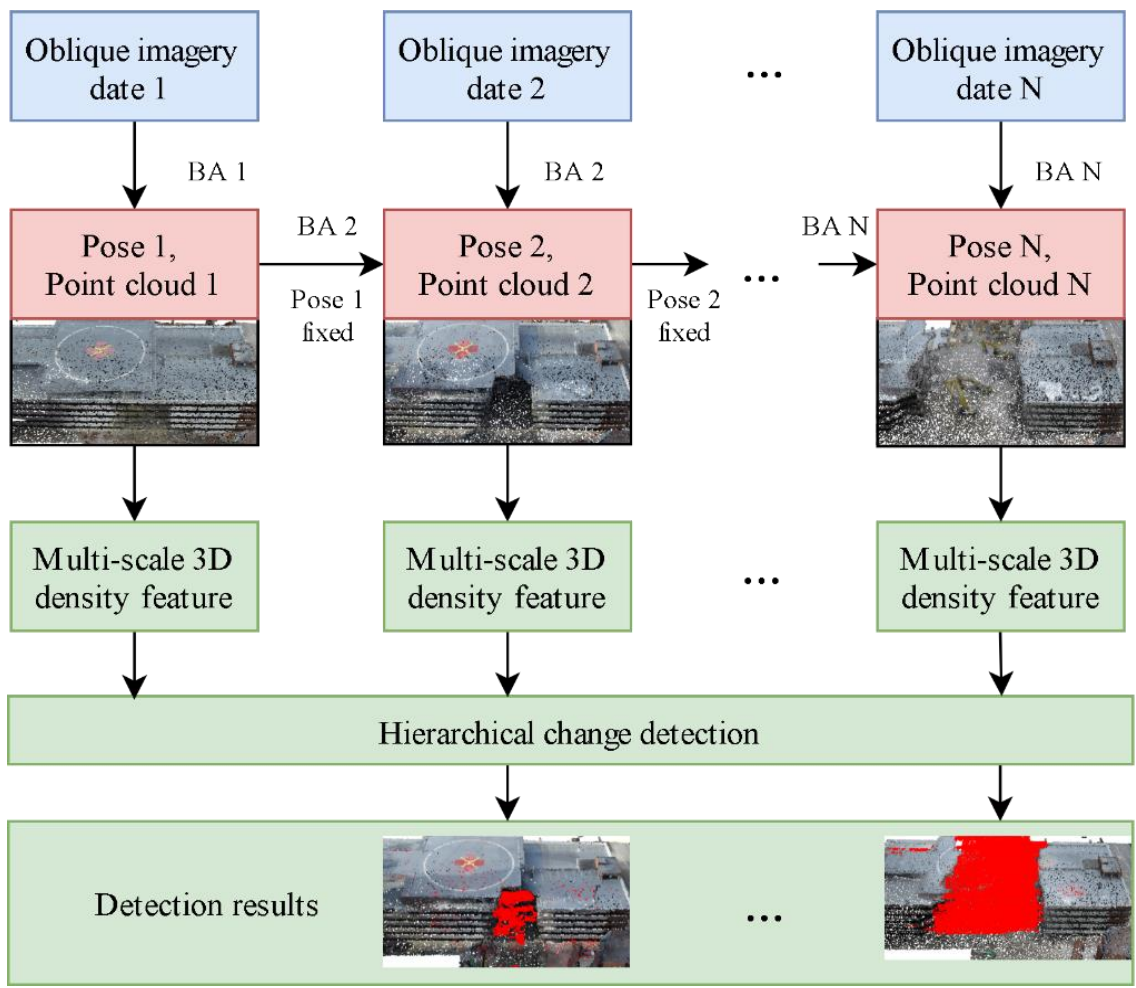

Figure 1 The general workflow of the proposed method. Details of this workflow are explained in the texts.

Figure 1 presents the proposed workflow, in which we start with date 1 oblique images (data collection of the test site introduced in Section 4.1), and perform a classic BA and dense point cloud generation through dense matching. The computed poses are used as fixed observations and fed into a successive BA for the oblique images captured from date 2, in which only the poses of the date 2 images are estimated. This is progressively performed as images of the new dates are collected (introduced in Section 3.1). The estimation of image poses from each date comes along with (multi-temporal) dense point clouds, of which we perform a hierarchical change detection algorithm using octree-based multi-scale 3D feature analysis (introduced in Section 3.2), and yield a changed area of each date. The proposed method aims to address two of the above-mentioned limitations when applied to our case study of demolition monitoring (details of this case study will be introduced in Section 4): 1) we build an implicit 3D registration paradigm by including progressive and time-dependent observations into UAV image geo-referencing, to improve the robustness and accuracy of 3D multi-temporal data alignment; 2) we propose to use an octree structure to represent the occupancy grid to reduce needed memory consumption and allow the flexibility to represent detailed structures, at the same time perform a coarse-to-fine detection to focus on major volumetric changes in 3D at a fine-grained level.

### *3.1 Progressive registration and pose-estimation of multi-temporal UAV imageries*

The proposed approach follows a feature-based implicit 3D registration paradigm as mentioned in Section 2.1, and the goal is to perform accurate pose-estimation across multi-temporal UAV-oblique image datasets through bundle adjustment (BA), such that the estimate poses for each of the images are consistent and within the same coordinate frame. The existing methods run a single BA across the multiple-temporal datasets, which estimates the pose of each image simultaneously (Li et al. 2017). As mentioned in Section 2.1, this introduces two potential drawbacks: 1) the number of tie points in a single BA grows exponentially leading to increased computations; 2) since the single BA covers images from the first to the last date, the scene contents might change dramatically leading to more outliers in tie points and unreliability in BA. We, hence, advance this paradigm with a minor modification to allow progressive BA by complying to two constraints: 1) the tie point matching is only performed between neighboring temporal datasets; 2) for each new temporal dataset $D_i$ , we only estimate the poses of images in $D_i$ while keeping the poses in the previous temporal datasets and observations $(D_{i-1}, ... D_1)$ fixed to resolve the datum for the new temporal dataset for pose estimation, at the same time to remove outliers of the newly detected tie points. The process is shown in Figure 1. To start, the first-date dataset is firstly geo-referenced via free-network BA, with datum decided by keeping the center of mass of the perspective centers the same as GPS positions. In addition, as a standard process, the GPS information will provide neighborhood connectivity to guide the tie point matching to reduce the computational cost. With this simple expansion of the previous approaches (Qin 2014; Li et al. 2017), the linearized observation formulation with a slight modification of the general BA procedure (Yao, Qin, and Chen 2019) is as follows Equation 1:

$$
\begin{aligned}
-e_B &= A_1 X_p + A_2 t_i + A_3 z_i + A_4 t_{i+1} + A_5 z_{i+1} - l_B : P_B \\
-e_t &= I_{t_i} - l_{t_i} : P_{t_i} \\
-e_z &= I_{z_i} - l_{z_i} : P_{z_i}
\end{aligned}
\tag{1}
$$

where $t_i$ and $z_i$ are the exterior orientation parameter (EOP) vector and self-calibration parameter (SCP) vector, respectively, for the georeferenced date $i$ as the reference here while $t_{i+1}, z_{i+1}$ are for date $i + 1$. $X_p$ denotes the object point vector recovered from date $i$ and date $i + 1$ , with $A_1, A_2, A_3, A_4$ and $A_5$ being their associated design matrix, respectively. $e_B$, $e_t$ and $e_z$ represent the true error vectors of image coordinates of all images on both date $i$ and date $i + 1$ , EOP, and SCP of the reference date $i$ . $l_B$, $l_{t_i}$ and $l_{z_i}$ are associated observation vectors of tie points between the two datasets, EOP and SCP of the reference date. $P_B$, $P_{t_i}$ and $P_{z_i}$ are the prior weights of the observations. EOP and SCP of the reference date $i$ are fixed as observations, thus $P_{t_i}, P_{z_i} \rightarrow \infty$.

The dense matching is performed using Agisoft Metashape (*Agisoft Metashape User Manual Professional Edition* (version Version 1.7) 2021) based on the optimized poses mentioned above separately for each date, to yield multi-temporal 3D model/point clouds. We evaluate the relative differences of the reconstructed surface at different dates to assess the level of 3D registration, a profile analysis of an unchanged surface is shown respectively in Figure 2, which has shown that surface collected from different dates coincide well and only minor differences are observed. The profile from the date "Jan. 14" appears to be slightly off from the other profiles, while still resides within a reasonable range (ca. ±5 cm). Table 1 evaluates the 3D geometric differences in the unchanged area of the entire dataset by taking the first-date dataset as the reference and shows that the mean registration accuracy ranges from 2-5 cm for a 2 cm image resolution, which has satisfactorily achieved an accuracy of 1-2 pixels.

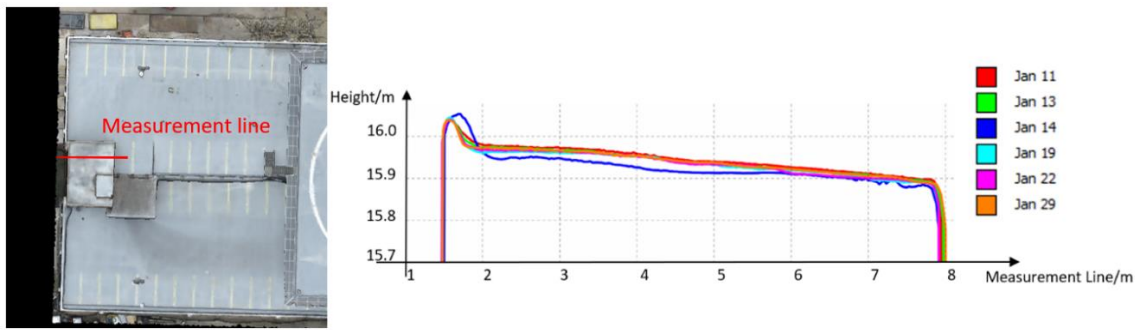

Figure 2 Profile analysis of an unchanged portion of surface throughout different dates showing the level of registration accuracy (statistics shown in Table 1).

Table 1 Registration accuracy of temporal dataset with respect to the reference day (January 11, 2021)

|  | Mean of distance (m) | Standard deviation of distance (m) |
| --- | --- | --- |
| January 13 | 0.0471 | 0.1495 |

| January 14 | 0.0339 | 0.0479 |
|---|---|---|
| January 19 | 0.0259 | 0.0302 |
| January 22 | 0.0513 | 0.0741 |
| January 29 | 0.0272 | 0.0362 |

### 3.2 Hierarchical volumetric change detection

With the produced dense point clouds from each date, we propose a hierarchical volumetric change detection method to compute the changes of multi-temporal point clouds. The idea is to utilize a coarse-to-fine strategy, to first locate areas where significant changes occur, and then progressively refine the changes on finer scales.

### 3.2.1 Multi-scale tree structure and hierarchical process

To set up a multi-scale structure, we used the octree (Shaffer 2011), a tree-based structure is used for partitioning the 3D space to adapt the distribution of the 3D point clouds. The octree tree starts its root note to represent the entire bounding box of the 3D point clouds and divided the space equally into eight volumes as child nodes (each is one-eighth of the parent node volume), and this division along depth (the number of divisions, or scales) is adaptively determined based on the density of the point clouds within each divided volume.

In this work, we divided the space by an octree with the depth of 11 as an empirical value leveraging the desired level of granularity and computation time, which gives to a volume of $1/8^{11}$ of the original volume as the finest scale (a volume of 0.027 m$^3$). The choice of depth can be also optionally determined by selecting the largest depth through portioning the point clouds given the desired density. The detection of changes on a certain scale can be achieved by calculating the distance of the 3D density feature per node (introduced in Section 3.2.2) for each node of the octree partition. Since the octree structure is adaptive with respect to the point clouds and varies with the point clouds, for this change detection purpose we used the octree structure generated by point clouds from date 1. Our hierarchical change detection process starts with a coarse level (a depth of 7 in our work). The detection on a finer scale is performed on the node (i.e. space represented by this node) detected as changes in its coarser scale. This effectively removes those isolated segments of changes while focusing on volumes of changes where major activities are. The process continues throughout the scales and a post-connected component analysis (Beksi and Papanikolopoulos 2016), as a common post noise-filtering method and easily extendable in 3D, is performed on the finest scale to obtain the final change detection results. Figure 3 shows an example of this process on the change detection of two dates, and points in each sub-figure (points of all colors) correspond to the detected changes on different scales, and the red portion remains the same throughout all the sub-figures, denoting the final detected results as a comparison. It can be seen that more detailed changes are detected as the maximal depth of oct-tree decomposition increases; large change segments are removed in the detections in a smaller depth (i.e. large granularity per node), thus avoid a large number of fractional noisy segments in the first place for the detection.

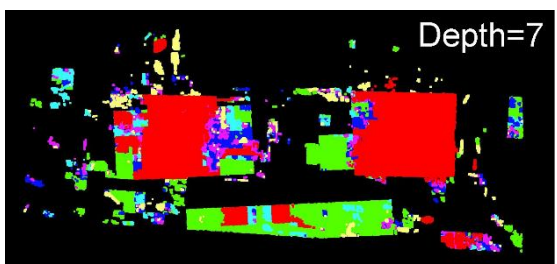

*(a) Detection result at depth=7*

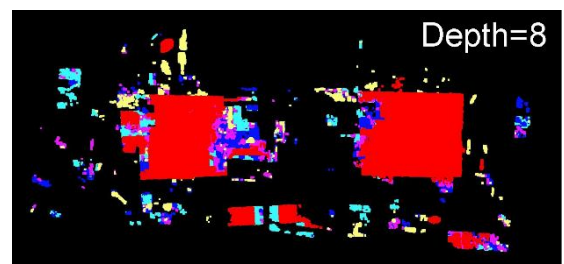

*(b) Detection result by removing green points from (a)*

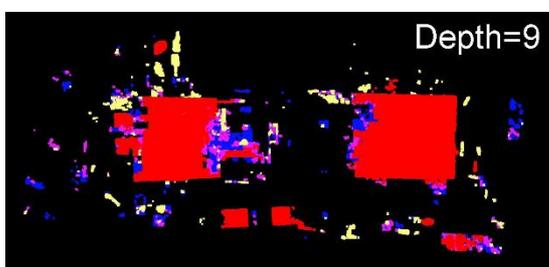

*(c) Detection result by removing light blue points from (b)*

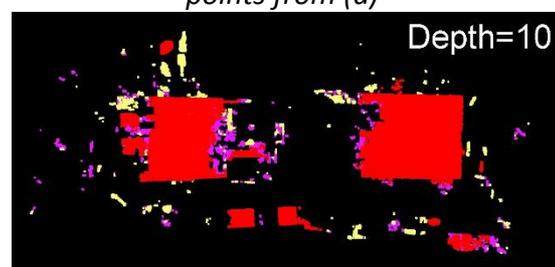

*(d) Detection result by removing blue points from (c)*

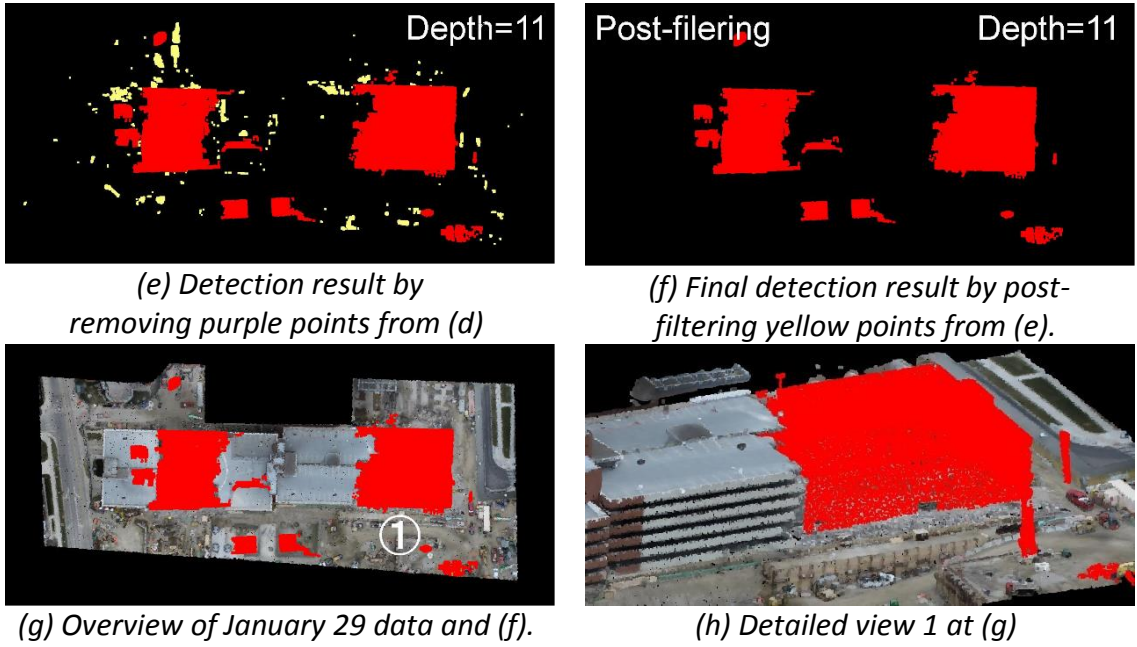

*(e) Detection result by removing purple points from (d)*

*(f) Final detection result by post-filtering yellow points from (e).*

*(g) Overview of January 29 data and (f).*

*(h) Detailed view 1 at (g)*

Figure 3 Hierarchical change detection process of point clouds of two dates (first and the last). "Depth" represents the scale level of the constructed octree. In each figure from (a) to (f), all points constitute the detection result at this depth. Red point cloud is the final detection result. Points of other colors are the points to be removed at different depths. "Post-filtering" in (f) was achieved by using connected components method.

### 3.2.2. 3D density feature for change vector analysis

Most of the existing methods perform an occupancy analysis for the volumetric space of the node under evaluation, which gives a binary description of change based on whether points are presented within the space for each point cloud. This can be particularly effective for point clouds with high fidelities such as LiDAR. To accommodate the potential uncertainties of image-based point clouds, we employ a "soft" approach by evaluating the possibility of changes within each node using a 3D density feature to form a change vector for analysis. For each volumetric space of the node under evaluation, we equally divide them into $N$ sub-voxels and the density of points of each sub-voxel can be concatenated and form a $N$ dimension feature vector $F$. Here the formed vector $F_1$ and $F_2$ for two point clouds are compared through their L-2 distance $D$ as an indicator for the possibility of changes, as

formulated as:

$$F_k = [d_{k,1}, d_{k,2}, \dots d_{k,i} \dots d_{k,N}], d_{1,i} \in R, k = 1,2$$
$$d_{k,i} = \frac{N_{k,i}}{V_{k,i}}, N_i, V_i \in R, k = 1,2 \tag{2}$$
$$D = \sum_1^N ||d_{2,i} - d_{1,i}||^2$$

where $N$ is the dimension of the feature vector, $D$ is the Euclidian distance between two feature vectors, $d_i$ is the density of a certain sub-voxel, $N_i$ is the number of points within the sub-voxel and $V_i$ is the volume of sub-voxels.

## 4. Experiments and Analysis

The UAV images were captured using a DJI Phantom 4 Pro model, and images are collected in six separate days at two adjacent garage structures at The Ohio State University campus on North Cannon Drive (in Figure 4). The buildings are 16 m in height and cover an area of 16,117 m$^2$. From January 11 to January 29, 2021 we captured 656, 547, 539, 638, 360, and 364 images respectively with nadir and façade viewing angles, and the collected dates do not cover the entire demolition process due to logistical constraints. The BA and dense point clouds generation are performed on a general workstation with an Intel Xeon W-2275 CPU, and dual GeForce RTX 2080 Ti graphics card. The processing time for each date data is approximately 90 minutes. The number of dense points of the generated 3D model ranges from 271,656 to 561,627 per date.

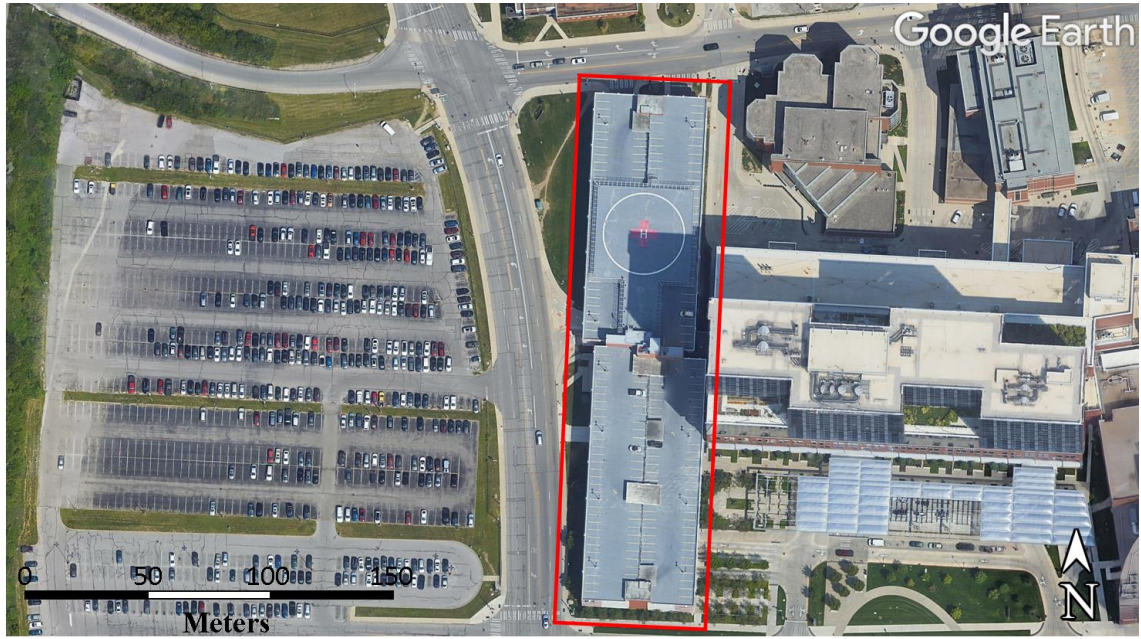

Figure 4 Study area is located at the North Cannon garage structures at The Ohio State University campus. Two adjacent buildings outlined in red are the object of interest that are subjected to demolition event during January-February of 2021.

## 4.1. Data collection

The oblique photogrammetric data are collected through five pre-defined flight trajectories, including a 45 degree off-nadir angle regular block (hereafter we call overview trajectory) with an 80% of both forward and side overlap. The overview flight design, in consideration of the camera focal length, has yielded a ground sampling distance (GSD) of approximately 2 cm on the roof of the buildings. The façades of the buildings are captured through four separate trajectories that follow the edge of the buildings to capture the details of the stories. In particular, the images are captured in a confined urban space where the UAV is flown close to the façade surface due to presence of high-rise nearby buildings (to the east of the buildings shown in Figure 4), which leads to a large scale difference with respect to images in the overview trajectory (with GSD varying from 0.7 cm to 2 cm). Therefore, the four façade trajectories are flown at two height layers such that the top layer serves as the convergence layer to be connected to images of the regular blocks, and particularly, to have the convergence angle less than 40 degrees with respect to the overview to allow successful tie point extraction. The sampled camera stations of each trajectory are shown in Figure 5, rendered through Pix4D software (*Pix4Dmapper 3.2 User Manual* (version 2017), n.d.).

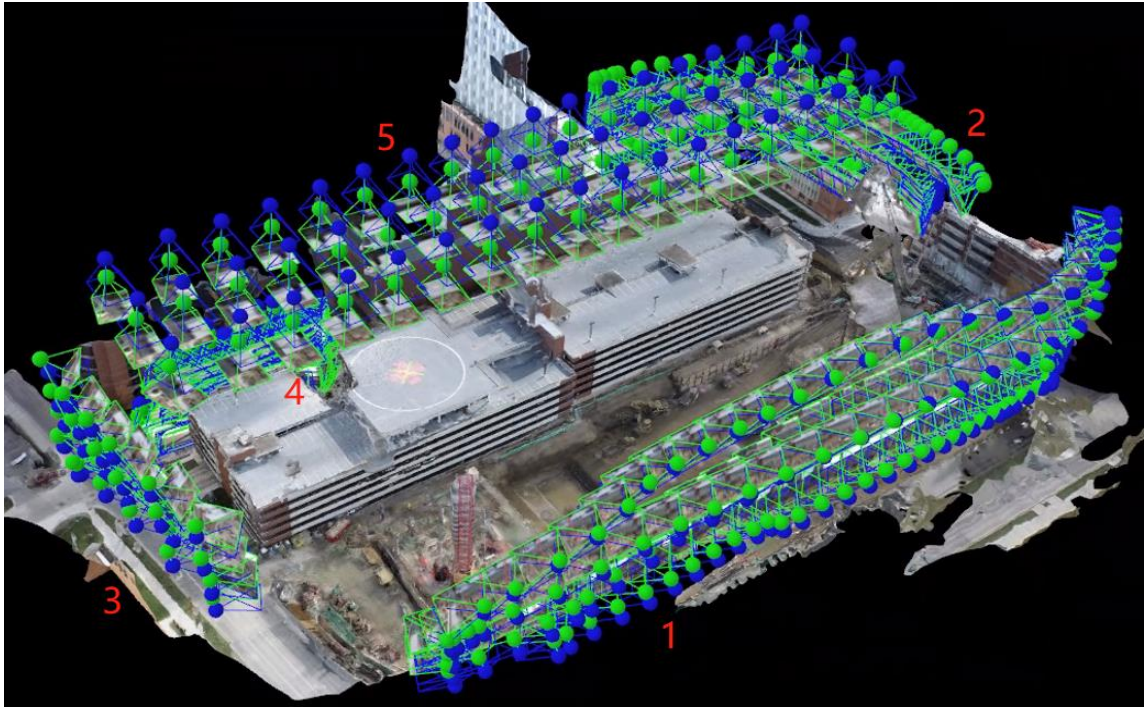

Figure 5 Flight trajectories of our monitoring mission with sequence numbers, visualization of the cameras produced by Pix4D software.

### 4.2 Accuracy analysis of the produced 3D point clouds from UAV images

To understand the metric accuracy of the 3D point clouds generated through UAV images, we compared them with the LiDAR point clouds collected both from an airborne platform (for top roof structure) through the State of Ohio and a Leica RTC360 terrestrial platform (for façade structure). We therefore respectively compute the point-to-plane distances between the drone data (from one date) and the LiDAR data for both roof surface and façade surface. For each point of the drone point cloud, the point-to-plane distance computes the Euclidean distance between the drone point and the plane formed by the closest $N$ points in LiDAR data.

Prior to the distance computation, the drone point clouds and the LiDAR point clouds are registered using the ICP (iterative closest point) algorithm (Besl and McKay 1992) to remove any systematic differences. The roof surface and one of the façade surfaces are extracted to measure the reconstruction accuracy and the results are shown in Figure 6 and Table 2. As shown in Table 2, we observed that in general, the drone-based point clouds in our acquisition configuration achieve 10-15 cm relative accuracy to LiDAR point clouds, and the accuracy for the roof structures appears to be slightly better than for the façade structure, even though the GSD of the façade is smaller than the roof structures. This attributes to two main reasons. First, the overview images capturing the roof geometry follow a well-designed photogrammetric block pattern, while the façade images have only a maximum of two strips. Second, the scene contents from the overhead perspective have richer contents and larger depth variation, both of which contribute to a more robust bundle adjustment and dense matching results.  As can be seen from bottom left illustration in Figure 6, the error map indicates that the eastern sides of the buildings have larger errors than the western sides, which is likely attributed to the suboptimal collection geometry of oblique images on the eastern side due to the nearby high-rise building (see Figure 4.). The accuracy for the façade as shown in Figure 6 (bottom right) shows no significant systematic patterns in the error maps although the overall accuracy is lower than that of the roof.

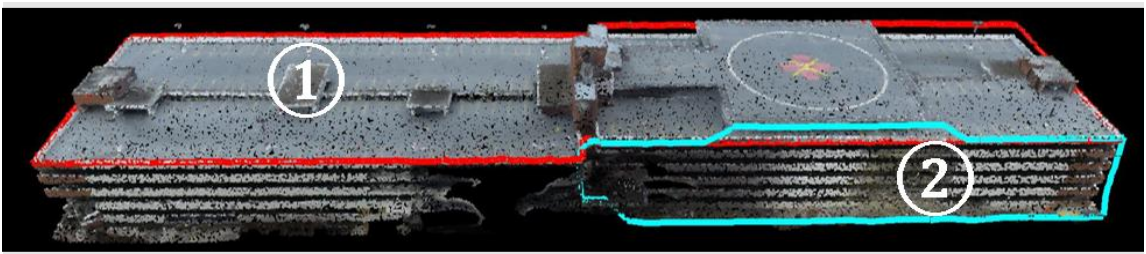

*Overview*

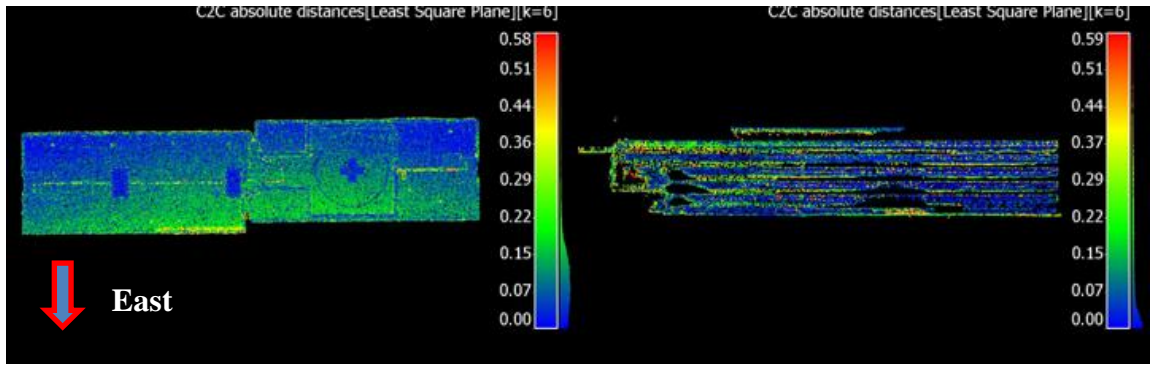

*(1) Roof surface*                              *(2) Façade surface*

Figure 6 Reconstruction accuracy for the roof surface and façade surface of drone data. (1) is the roof area of drone point cloud and its error map with respect to airborne LiDAR(through the State of Ohio) while (2) is the facade surface of drone point cloud and its error map with respect to LiDAR(through a Leica RTC360 laser scanner).

Table 2 The distance statistics of top-view and facade-view.

|                          | Roof surface | Façade surface |
|--------------------------|:------------:|:--------------:|
| Average distance (m)     | 0.1255       | 0.1651         |
| Standard deviation (m)   | 0.1088       | 0.1514         |

### 4.3. Change detection accuracy analysis

The change detection is performed for consecutive dates. To analyze the accuracy of the change detection accuracy, we quantitatively evaluate the results for the first date (January 11th) and the last date (January 29th) data. We selected four representative regions of changes in the building and manually labeled the changes to serve as the ground-truth for the assessment, respectively. Two surfaces are from the roof and two are from the façades, as shown in Figure 7 (first row). The detailed views of these four regions are in the second and third rows of Figure 7. These regions are selected to both include change and unchanged areas and our detection results are shown as true positives (TP, in green), false positives (FP, in blue) and false negatives (FN, in red). True negatives (TN) are pixels without any masks. In general, the detected changes coincide well with the ground truth as a result of the large proportion of the TPs, and it can be observed that the proposed method underestimates the changes given the notable number of FNs. This is counter-intuitive to most of the change detection studies reported in the literature since they tend to overestimate the changes given the uncertainty of the data for comparison and other disturbances factors such as moving or temporary objects. Since the proposed change detection application explores the full-scale 3D information with a robust density-based change vector, any subtle changes within a volumetric unit, if not significantly introduce density changes, will not be captured unless the depth of the octree is increased to have finer scales. This is expected in our specific application, due to the fact that changes occurring in a demolition site are rather focused, thus we aim our method to minimize potential false positives by robustly highlighting the focused changes. We expect that increasing the depth will reduce the FNs and may potentially introduce more FPs, and vice versa. Therefore, a depth of 11 in our experiment is an empirical trade-off that gives realistic volumetric change estimations, as the FN regions, although notable, only contains relatively minor changes in volume and will unlikely to pose significant impact in volume changes.

We use different metrics including intersection over union (IOU), precision, recall, and F1 score as the measures of accuracy. Results of these four regions are in Table 3, and it can be seen that our proposed method in general has yielded an F-1 score ranging from 0.78 to 0.92, indicating the level of agreement between the detection and the ground truth. Although the highest F1-score is achieved on the roof surface and the lowest on the façade, we do not observe a notable systematic difference in the accuracy achieved for roof and façade structures. In addition, our method has achieved consistently a high precision in detection, this is of particular use when being used for calculating volumes of changes.

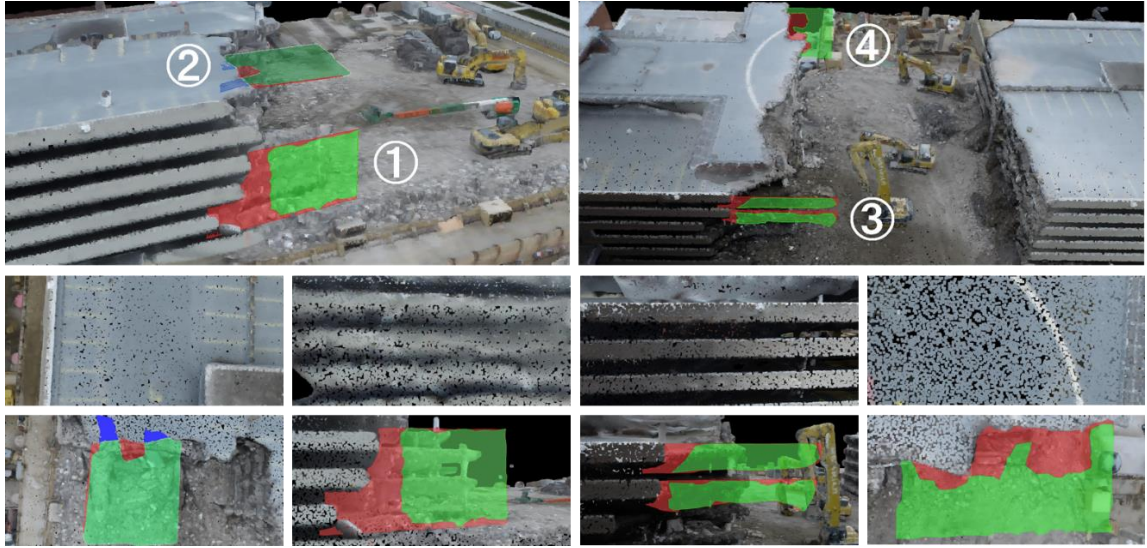

Figure 7 Change detection accuracy evaluation of date 1 (January11) and date 2 (January 29). The first row: overview of the four selected regions; The second and third rows: enlarged views of the results for the four regions. Green: true positives (FPs), Blue: false positives, Red: false negatives (FNs).

Table 3. Accuracy evaluation of the change detection for the selected regions

|  | IOU | Precision | Recall | F1-score |
|---|---|---|---|---|
| Roof 1 | 0.8563 | 0.9299 | 0.9153 | 0.9225 |
| Roof 2 | 0.6878 | 0.9982 | 0.6886 | 0.8150 |
| Façade 1 | 0.7858 | 0.9967 | 0.8801 | 0.8801 |
| Façade 2 | 0.6440 | 0.9999 | 0.6440 | 0.7834 |

## 4.4. Derived volumetric changes

By progressively applying the proposed methods on the data of these six dates (following the general workflow in Figure 1), we visualize the detected changes in point clouds showing the building demolition processes, colorized based on the date of changes in Figure 8, where the arrow shows the progress where the demolition starts and moves towards the other parts of the buildings. Demolition starts from the middle of the building in the North (Figure 4) and it starts from the edge in the South building.

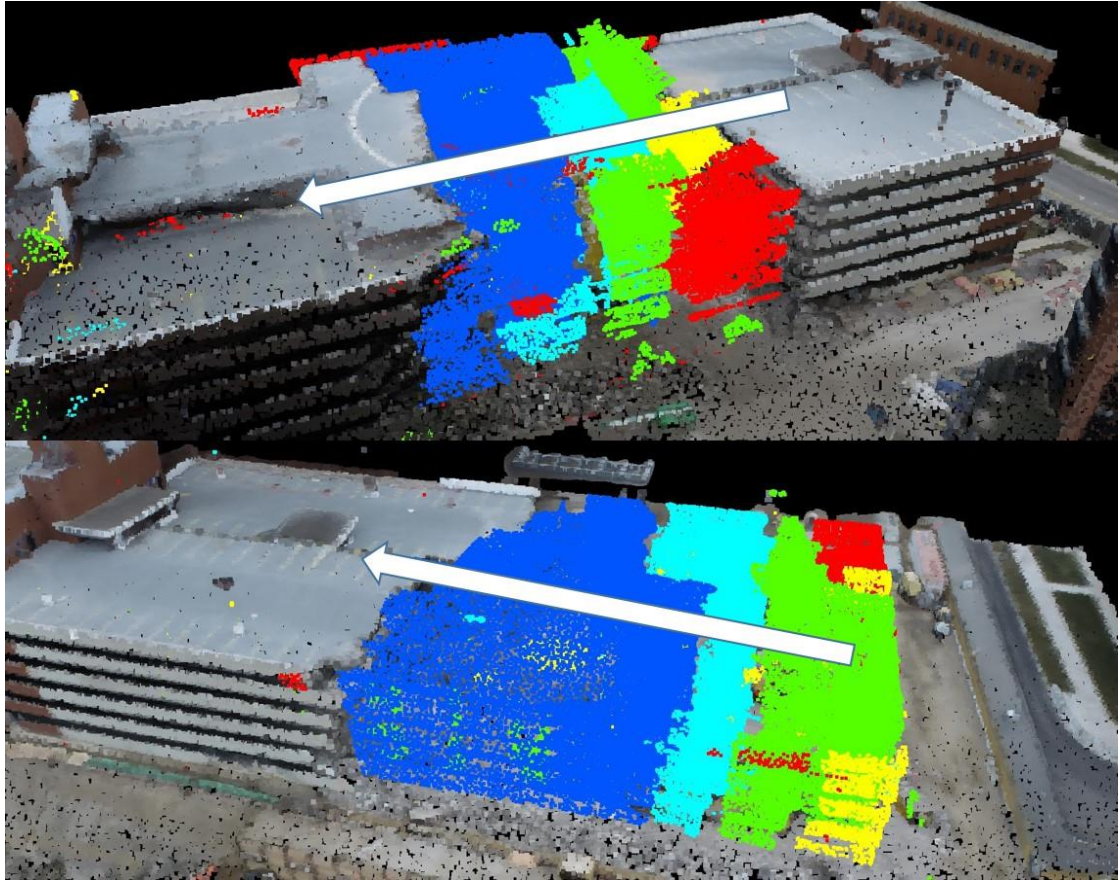

Figure 8 Visualization of the building demolition process. The arrow indicates the directions of progression of demolition. Portion being demolished in different dates are colorized in Red (January 11-January 13), Yellow (January 13-January 14), Green (January 14-January 19), Light Blue (January 19-January 22), and Blue (January 22-January 29).

We further compute the volumetric changes throughout the demolition process. To compute, the point cloud, partial meshes need to be converted to a complete and watertight surface model. However due to the often-unavoidable occlusions in data collection, directly taking the interpolated surface model will result in significant overestimation of volume, especially for multi-story buildings. We, in turn, utilize the octree structures that we introduced before for each pair of temporal point clouds. Then, we compute the changed volume within the cubic volumetric space (a voxel) of the finest scale, which is detailed as: for two point clouds $P$ and $Q$, an octree is built for one of the point clouds where the changing voxels $K$ on the finest scale can be identified. The changed voxels $K$ are projected to the ground plane, forming a 2.5D grid map. The changing volume of each occupied grid can be regarded as a mini-2.5D surface. The volume of changes is then computed as the grid area times the height difference of $P$ and $Q$, within that grid. The changing volume $V$ is computed by summing up the changing volume of all occupied grids:

$$V = \sum_i s^2 \cdot h_i \qquad (3)$$

where $V$ is the volume, $i$ is the index of the occupied grid, $s$ is the grid size, $h_i$ is the difference of height in the grid $i$. Figure 9 depicts the cumulative volume of change and the rate of the changes during the demolition process, and it can be seen that from date 1 to date 6, there are in total 40,481.8 m³ volume of masses demolished. The speed of demolition starts slow, at a rate of 967.3 m³/day and keeps relatively stable (varying from 1152.8 m³ / day to 1774.2 m³ / day), and almost doubled at the last seven days (2479.5 m³ / day). This as a result leads to a 19836.1 m³ of demolitions in the last seven days, as compared to 20645.7 m³ since the beginning of the demolition to the beginning of the last week.

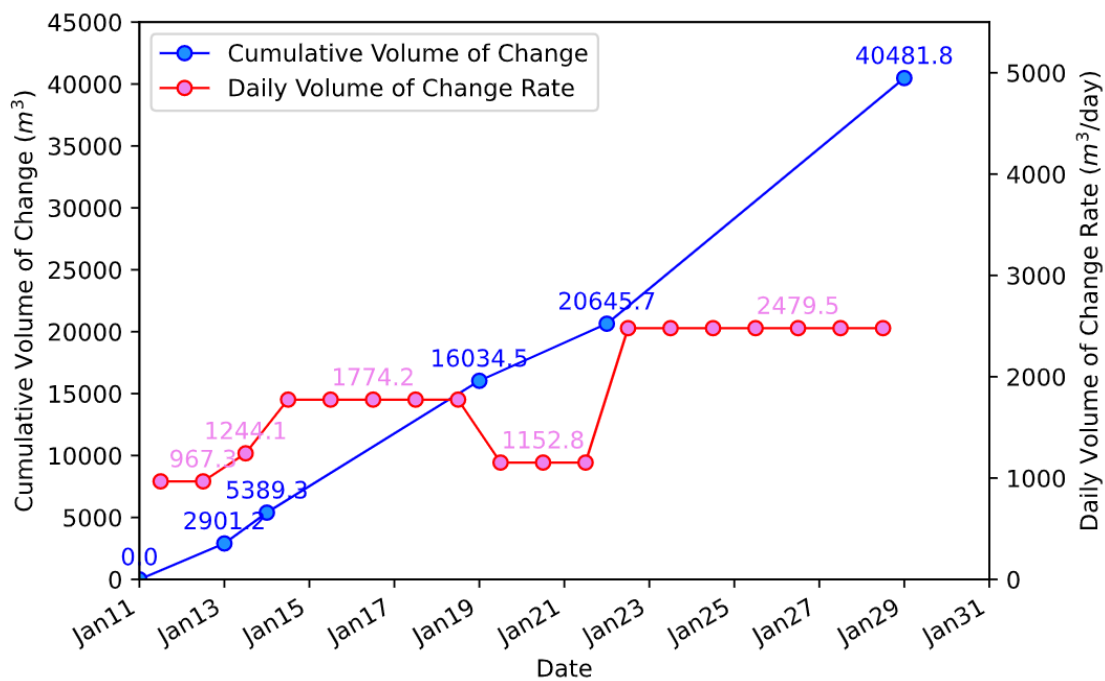

Figure 9 Cumulative volume change and daily volume change rate detected using the proposed framework.

## 5.Conclusions

In this paper, we presented an end-to-end workflow for 3D change detection using time-series UAV-oblique images. The idea is to perform a fully automated progressive bundle adjustment and a coarse-to-fine 3D volumetric change detection framework to locate and identify the volume of changes. The proposed progressive bundle adjustment follows and advances existing 3D implicit registration paradigms to allow 3D collections to be robustly registered for evaluation, and the octree-based volumetric change detection algorithm with a 3D density change vector is specifically designed to reduce false positives commonly presented in change detection to ensure high precision.

The proposed workflow is validated on a case study of building demolition event through progressive collapse, in which two adjacent five-story parking garage structures (with approximately 35,000 square meters of footage) underwent a demolition process and UAV-oblique photogrammetric images are collected under a constrained collection environment, repetitively on six separate days throughout the one-month demolition period. Our work has validated the proposed approach through the following three aspects:

(1) We evaluated the generated UAV point clouds through cloud-to-cloud distance with airborne and terrestrial LiDAR point clouds and concluded that our collection configuration (as introduced in Section 4.1) yields point clouds with an accuracy of 12 cm on the roof structure and 16 cm on the façade structure.

(2) We evaluated the proposed hierarchical change detection algorithm against the selected regions where ground-truth changes are manually labeled and achieved an F-1 score varying from 0.78 to 0.92, with consistently high precision $(0.92 - 0.99)$, which is suitable to identify focused changes occurring in a progressive manner.

(3) We calculated the volumetric changes using the collected data and demonstrated that the derived statistics such as the total volume of change and the demolition rate may serve as useful information for construction management or for assessment of structures that may be damaged or partially collapsed after a man-made or natural disaster.

We consider this work has presented a useful workflow and a full-scale case study that will provide useful knowledge to potential researchers and engineers for such ultra-high-resolution monitoring tasks using UAV-oblique photogrammetry. In our work, we noted the drawbacks of this method still lies in the lack of control of octree depth, as well as a more accurate presentation for volumetric calculation under cases where partial occlusion of the geometry exists Therefore, our future work will attempt to address these challenges.

**Acknowledgement:**

The authors would like to thank the State of Ohio to make the airborne LiDAR available through Ohio Geographically Referenced Information Program and Woolpert. Inc to scan the building using their terrestrial laser scanner.

**Funding**

This material is based upon work supported by the U.S. National Science Foundation [grant number 2036193]. The authors are supported in part by Office of Naval Research [grant number N00014-17-l-2928], [grant number N00014-20-1-2141].

**Data Availability Statement**

The data associated with this paper are produced by the authors; for any publicly available data, the authors have acknowledged the use of them in the **Acknowledgement** section.